\documentclass[5p,twocolumn,10pt,times]{elsarticle}
\usepackage{amsmath}
\usepackage{hyperref}
\usepackage{graphicx}
\usepackage{multirow}
\usepackage{algorithm}
\usepackage{algpseudocode}
\usepackage{comment}
\usepackage{xcolor}
\usepackage[normalem]{ulem}
\newcommand{\TM}[1]{{\textcolor{black}{#1}}}

\newcommand{\YXL}[1]{{\textcolor{black}{#1}}}
\newcommand{\LXY}[1]{{\textcolor{black}{#1}}}
\begin{document}
\baselineskip11pt

\begin{frontmatter}

\title{RLCAD: Reinforcement Learning Training Gym for Revolution Involved CAD Command Sequence Generation}

\author[1,†]{Xiaolong Yin} 
\author[1,†]{Xingyu Lu} 
\author[1]{Jiahang Shen}
\author[1]{Jingzhe Ni}
\author[2]{Hailong Li}
\author[1]{Ruofeng Tong}
\author[1]{Min Tang}
\author[1]{Peng Du\corref{cor1}}
\fntext[†]{Equal contribution} 
\cortext[cor1]{Corresponding author}
\ead{dp@zju.edu.cn}
\address[1]{Zhejiang University, China}
\address[2]{Shenzhen Poisson Software Co., Ltd., China}

\begin{abstract} 
A CAD command sequence is a typical parametric design paradigm in 3D CAD systems where a model is constructed by overlaying 2D sketches with operations such as extrusion, revolution, and Boolean operations. Although there is growing academic interest in the automatic generation of command sequences, existing methods and datasets only support operations such as 2D sketching, extrusion, and Boolean operations. This limitation makes it challenging to represent more complex geometries.

In this paper, we present a reinforcement learning (RL) training gym specifically designed for CAD model generation, along with an RL-based algorithm that generates command sequences from boundary representation (B-Rep) geometry within this training gym. Given an input B-Rep, the policy network of the RL algorithm first outputs an action. This action, together with previously generated actions, is processed within the gym to produce the corresponding CAD geometry, which is then fed back into the policy network.
Rewards, computed by the difference between the generated and target geometries within the gym, are used to update the RL network. Our method supports operations beyond sketches, Boolean, and extrusion, including revolution operations. With this training gym, we achieve state-of-the-art (SOTA) quality in generating command sequences from B-Rep geometries. 

\end{abstract}


\end{frontmatter}


\section{Introduction}
\label{sec:introduction}
\begin{figure*}[tb]
  \centering
  \includegraphics[width=1.0\textwidth]{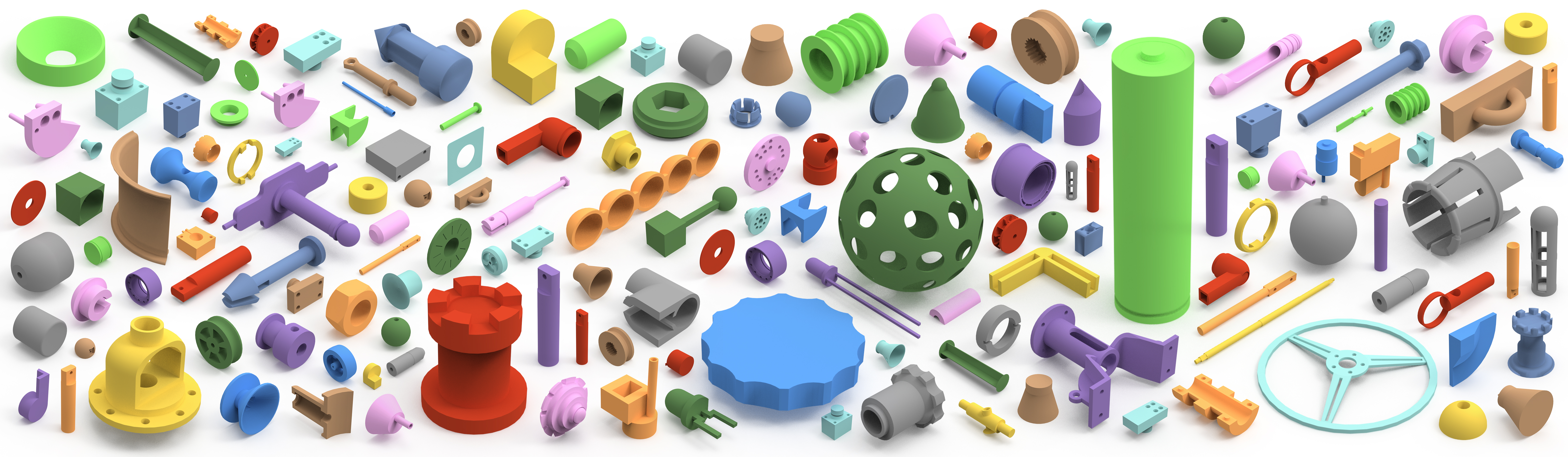} 
  \caption{
  Demonstration of various CAD models generated by our novel CAD model generation algorithm. Including revolution operation enables the generation of more complex geometric shapes, such as the battery, nuts, flanges, etc.
  } 
  \label{fig:dataset}
\end{figure*}

In our daily lives and industrial production, objects ranging from small items like cups to large structures like airplanes are designed using a series of 2D and 3D modeling operations in computer-aided design (CAD) software. The process of recording these parametric operations results in a command sequence that encapsulates domain expert knowledge by precisely defining a sequence of modeling steps to generate CAD geometries. However, obtaining the command sequence directly is often not feasible. Instead, we should generate the corresponding command sequence using other modalities, such as boundary representation, meshes, or point clouds.

Over the past few years, generative models have flourished, demonstrating immense potential. Representative approaches include diffusion models~\cite{dhariwal2021diffusion}, GANs~\cite{mirza2014conditional}, autoencoders~\cite{kingma2013auto}, large language models (LLMs)~\cite{vaswani2017attention,devlin2018bert}, and vision-language models (VLMs)~\cite{chen2022pali}. These methods have been extensively applied in generating CAD models, such as boundary representation (B-Rep) autoregressive encoder BrepGen~\cite{xu2024brepgen}, end-to-end point cloud-to-command sequence generation model CADSIGNet~\cite{khan2024cad}, text-to-command sequence generation model Text2CAD~\cite{khan2024text2cad}, multi-modal command sequence generation model CAD-MLLM~\cite{xu2024cad}, B-Rep to command sequence generation model Fusion 360 Gallery~\cite{willis2021fusion}, voxel-to-command sequence generation model SECAD-Net~\cite{li2023secad} and SfmCAD~\cite{li2024sfmcad}, autoregressive command sequence generation models HNC-CAD~\cite{xu2023hierarchical}, SkexGen~\cite{xu2022skexgen}, and DeepCAD~\cite{wu2021deepcad}, as well as LLM based CAD code generation algorithm~\cite{alrashedy2024generating} and platform Zoo.dev~\cite{zoo_dev}. Despite their success, most of these methods support only basic operations like 2D sketch, extrusion, and Boolean operations.

Traditional CAD geometry generation algorithms rely on CAD geometric engines, which provide essential functionalities such as validity checks and constraint solving -- capabilities that supervised learning alone cannot achieve. Meanwhile, SOTA large language models such as DeepSeek R1~\cite{guo2025deepseek} and OpenAI o1~\cite{jaech2024openai} employ reinforcement learning (RL) techniques to enhance text generation quality and model reasoning capabilities. In bionic robotics, RL-based simulation training has also yielded impressive results in robot motion control~\cite{zhao2024zsl} and manipulation~\cite{fu2024mobile}. 

\YXL{Inspired by these advancements, we present the first RL training gym tailored to parametric CAD command sequence generation on exact B-Rep geometry. We provide a high-level Python interface wrapper and support parallel environment sampling, enabling efficient interaction with multiple CAD environments. In this framework, first we represent the B-Rep as a face-adjacency graph, then employ a graph neural network based feature extractor to encode the B-Rep, and finally use a policy network to predict an action. Each action is described by four elements—the start face, end face, operation type (newbody, intersection, union, or subtraction), and action type (extrude or revolve), where the start and end faces differ for extrusion but coincide for revolution—which is executed in the gym to produce the corresponding CAD geometry. The resulting geometry is then fed back into the policy network to compute rewards and update the network parameters.}

\YXL{
Our approach supports operations beyond sketches, Boolean, and extrusion, including revolution operations. As shown in Figure~\ref{fig:dataset}, including revolution operation enables the generation of more complex geometric shapes, such as the battery, nuts, chess pieces, etc.} 


\YXL{\paragraph{\bf Technical contributions} We present a CAD-focused RL training gym and an RL-driven CAD model generation algorithm that converts B-Rep models into command sequences within the training gym. Our approach features a multi-modal policy network that integrates multiple common CAD operations with cross-modal feature extractors. To further enhance generation quality, we introduce a set of hybrid reward functions, including geometric similarity and neural similarity, which jointly guide the learning process and help better distinguish models in both geometric and semantic aspects.}


\TM{We construct an RL-oriented dataset based on the ABC dataset~\cite{koch2019abc}, applying filtering and de-duplication to obtain a curated set of $20k$ CAD models that contain both extrusion and revolution operations, together with their ground-truth command sequences. Unlike existing extrusion-dominated CAD benchmarks, our dataset expands the operational repertoire and is intended to support future research on general CAD command-sequence generation.}



\section{Related Work}
\label{relatedwork}
We review related work in four aspects: CAD datasets, CAD model generation, B-Rep to command sequence generation, and RL training gyms.

\subsection{CAD Datasets}
Current parametric CAD models typically are stored as B-Rep~\cite {braid1975synthesis} or command sequence~\cite{price1982design}. B-Rep includes the analytical representation of the surface and curves of a 3D shape, while command sequences record the historical construction process of the shape. B-Rep models consist of trimmed parametric surfaces along with topological information describing their adjacency relationships.

The ABC dataset~\cite{koch2019abc} is the most widely used B-Rep dataset in academia, containing one million CAD models. DeepCAD~\cite{wu2021deepcad} cleaned 170,000 pairs of B-Rep and command sequences from the ABC dataset. Text2CAD~\cite{khan2024text2cad} constructed hierarchical descriptions for the DeepCAD dataset, ranging from coarse to fine granularity, using LLMs and VLMs, adding text descriptions to the command sequences. CAD-MLLM~\cite{xu2024cad}, using a method similar to DeepCAD, extracted 185,000 models from the ABC dataset and generated corresponding text descriptions, multi-view images, and point clouds for these models. However, these datasets primarily contain 2D sketches, such as lines, arcs, and circles, along with extrusion and Boolean operations. 

\subsection{CAD Model Generation}
CAD model generation involves generating a B-Rep model or command sequence from inputs such as B-Rep models, multi-view images, textual descriptions, or point clouds. CAD reconstruction remains a challenging problem because it requires reasoning over continuous parameters (e.g., dimensions of basic solids) and discrete operations (e.g., Boolean and extrusion operations), resulting in a hybrid combinatorial search space.

BrepGen~\cite{xu2024brepgen} combines vector variational autoencoders and latent space diffusion models for the autoregressive generation of B-Rep models. CADSIGNet~\cite{khan2024cad} employs layer-wise cross-attention between point clouds and command sequences to learn implicit representations of CAD models, enabling end-to-end point cloud to command sequence generation. HNC-CAD~\cite{xu2023hierarchical}, SkexGen~\cite{xu2022skexgen}, and DeepCAD~\cite{wu2021deepcad} quantize and encode CAD command sequences, generating corresponding latent space features and training through an encoder-decoder autoregressive model. SfmCAD~\cite{li2024sfmcad} uses voxels as input and proposes differentiable sketch and sweeping path modeling operations to reconstruct the CAD model in a self-supervised manner.  

\subsection{B-Rep to Command Sequence Generation}
Several commercial CAD softwares~\cite{solidworks,inventor} employ heuristic feature extraction to generate command sequences from B-Rep models. However, the 2D sketches extracted using this method are often structurally complex and difficult to edit further. Fusion 360 Gallery~\cite{willis2021fusion} is a representative deep learning-based approach that first extracts features to obtain continuous geometric parameters for basic operations (e.g., 2D sketching, extrusion, etc.). It then uses imitation learning to generate confidence scores of the command sequence from the input B-Rep model. Finally, a confidence-based local search algorithm, combined with geometric similarity comparison, optimizes model accuracy in a simulation environment. However, this method only supports face-extrusion operations, and the local search algorithm heavily relies on the accuracy of the confidence scores from supervised learning.

\subsection{RL Gym} 
RL involves updating states and receiving rewards or penalties through agent-environment interactions to maximize long-term returns, thereby learning decision-making strategies~\cite{wiering2012reinforcement}. Based on environmental modeling methods, RL can be divided into model-based and model-free categories. 

Model-based approaches first model the state transition function and reward function of the environment using small-scale datasets, then optimize the policy network under the RL framework using the environment model~\cite{moerland2023model}. A representative method is RL with Human Feedback~\cite{christiano2017deep}, widely used in pre-training large language models. Model-free methods replace the environment model with real-world environments or physical simulators, obtaining immediate feedback on the policy network through real-time interaction within the RL framework~\cite{ccalicsir2019model}. These methods are widely applied in bionic robot training simulations, such as robot motion control and dexterous hand grasping~\cite{zhao2024zsl,fu2024mobile}. Representative robot simulation engines include Isaac~\cite{makoviychuk2021isaac}, MuJoCo~\cite{todorov2012mujoco}, and Genesis~\cite{genesis}.

Therefore, we propose to build a CAD training gym based on the CAD geometric engine, Parasolid~\cite{ParasolidWikipedia}, which can be easily extended to integrate with other CAD engines, such as OpenCascade~\cite{opencascade} or ACIS~\cite{acis}.

\begin{figure*}[t]
  \centering
  \includegraphics[width=1.0\textwidth]{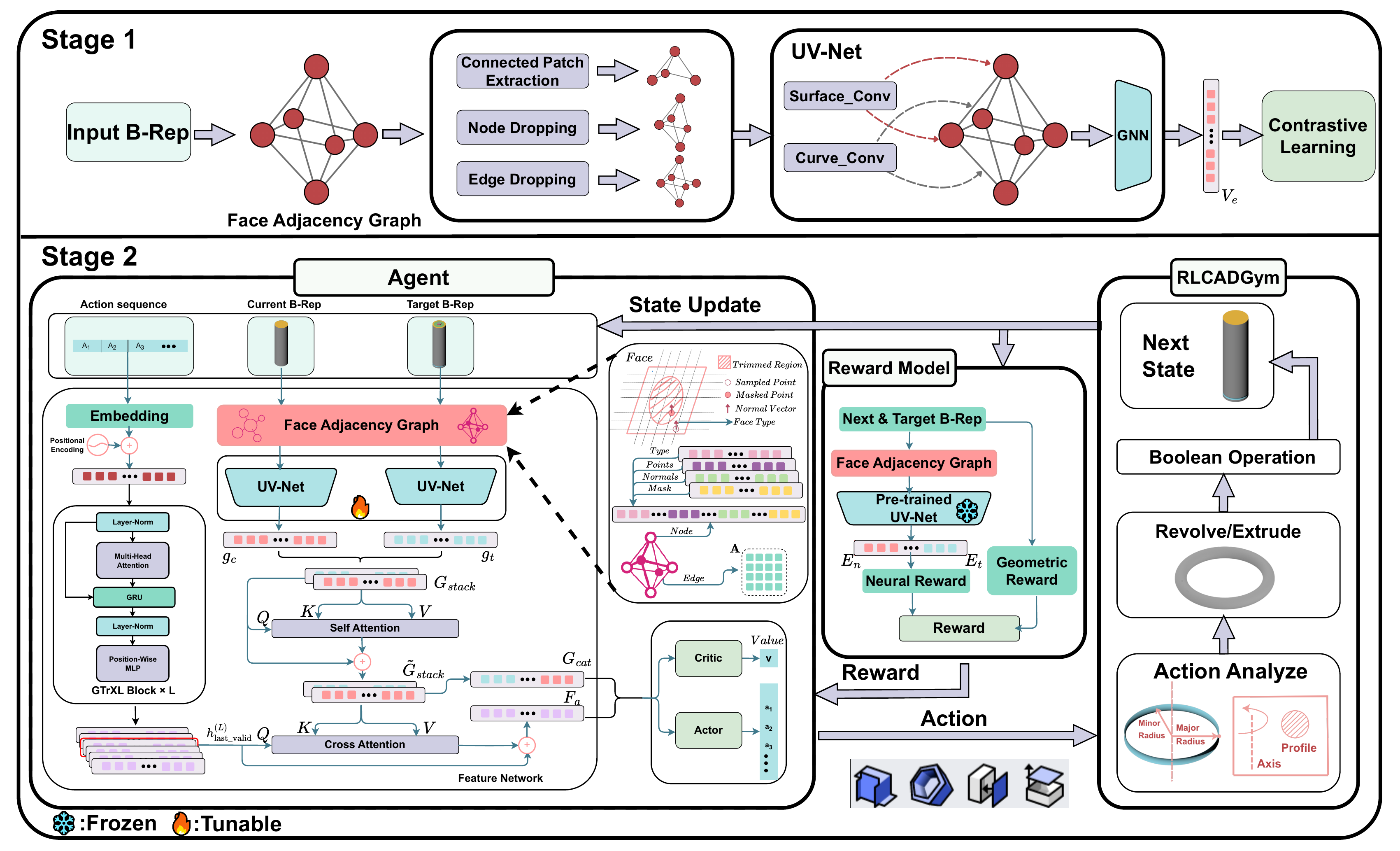} 
   \caption{The training pipeline is composed of two stages. In the first stage, a contrastive learning approach is employed to pre-train the UV-Net network, aiming to derive an encoder model that can effectively characterize the B-Rep of the CAD model. During the second stage, a reinforcement learning approach is employed to generate the command sequence. We first utilize the tunable UV-Net model to extract the B-Rep embedding of the CAD model, which is then integrated with the feature vector of the historical modeling action sequence. Subsequently, the Actor-Critic network predicts the action distribution and value. The predicted action is transmitted to RLCADGym for execution, yielding the next-stage observation. The neural reward and geometric reward are designed to update the policy network.}
  \label{fig:pipeline} 
\end{figure*}

\section{Overview and Preliminary}
\label{sec:overview}

In this section, we provide an overview of our approach and introduce key preliminaries. 

\subsection{Our Approach}

RL is an unsupervised method where an agent achieves its goals through interaction with an environment. The agent outputs its actions to the environment, and the environment, upon receiving these actions, proceeds to the next step, returning the next observation and the reward generated by the action to the agent. This interaction produces a series of observations, and the agent's objective is to learn a policy that maximizes the cumulative reward from these observations. 

The overall training pipeline is depicted in Figure~\ref{fig:pipeline}. In the first stage, we adopt a contrastive learning approach to pre-train UV-Net~\cite{jayaraman2021uv} on a large-scale dataset of unlabeled CAD models, aiming to derive an embedding that can effectively characterize the high-dimensional B-Rep feature of CAD models. Given a B-Rep model represented as a face adjacency graph, UV-Net processes it through a combination of surface convolution and curve convolution operations, followed by a Graph Neural Network (GNN) to extract a compact and discriminative embedding. To enhance the robustness of the learned representations, we apply UV-Net's data augmentation strategies, including connected patch extraction and stochastic node/edge dropping.


In the second training stage, the whole network can be divided into two parts: the feature extraction network and the Actor-Critic network. Inspired by advancements in natural language processing (NLP), we consider incorporating the action sequence as part of the state and fusing it with B-Rep features using cross-attention. This allows us to measure the gap between the current action sequence and the target geometry. Specifically, for the current B-Rep and target B-Rep, we use the pre-trained UV-Net to extract features from the face adjacency graphs, obtaining \( g_c \) and \( g_t \). These features are stacked to form \( G_{\text{stack}} \), and self-attention is applied to fuse \( g_c \) and \( g_t \), resulting in \( \tilde{G}_{\text{stack}} \). For the action sequence, we use an \( L \)-layer GTrXL network to extract features. And then we take the last valid action \( h_{\text{last\_valid}} \) as the feature of the action sequence. Cross-attention is then applied to fuse \( h_{\text{last\_valid}} \) and \(\tilde{G}_{\text{stack}}\). Finally, the fused feature serves as input to the Actor and Critic networks, predicting the probability distribution of actions and the cumulative reward of the current state, respectively. The action is sampled from the probability distribution and interacts with RLCADGym to obtain the reward and the next state. Through multiple rounds of interaction and network parameters update, the optimal action sequence is predicted.


\subsection{Preliminary}
We formulate the CAD reconstruction task from B-Rep to a command sequence as a Markov Decision Process (MDP). The key components of this formulation are defined as follows:
States $(s)\in\mathcal{S}$ contains the current and target B-Rep geometry, we use a face-adjacency graph~\cite{jayaraman2021uv} to represent the B-Rep geometry. Action $(a)\in\mathcal{A}$ represents a modeling operation that allows the simulator to modify the current geometry with extrusion, revolution, and Boolean operations. State transition kernel $T$ allows the simulator to apply the modeling operation to update the current geometry. Reward $(r)\in\mathcal{R}:\mathcal{S}\times\mathcal{A}\to\mathbb{R}$ allows the simulator to calculate the difference between the current and target states, and reward discount $\gamma\in[0,1)$,  determining the importance of future rewards.

Our objective is to optimize a policy network $\pi_\theta(a_t|s_t)$ that governs the selection of modeling operations. The episodic trajectory $\tau$ generated by following policy $\pi_\theta$ is given by: 
\begin{equation*}
    \tau=\Big( (s_0,a_0,r_0), ..., (s_{|\tau|},a_{|\tau|},r_{|\tau|})\Big).
\end{equation*}
The RL objective is to maximize the expected cumulative rewards: 
\begin{equation*}
    \text{maximize} ~ \mathbb{E}_{\tau\sim\pi_\theta}\bigg[ \sum_{t=0}^{|\tau|}\gamma^{t}r_t \bigg].
\end{equation*}

\section{CAD Training Gym}
\label{sec:gym}

\begin{figure*}[tb]
  \centering
  \includegraphics[width=0.9\textwidth]{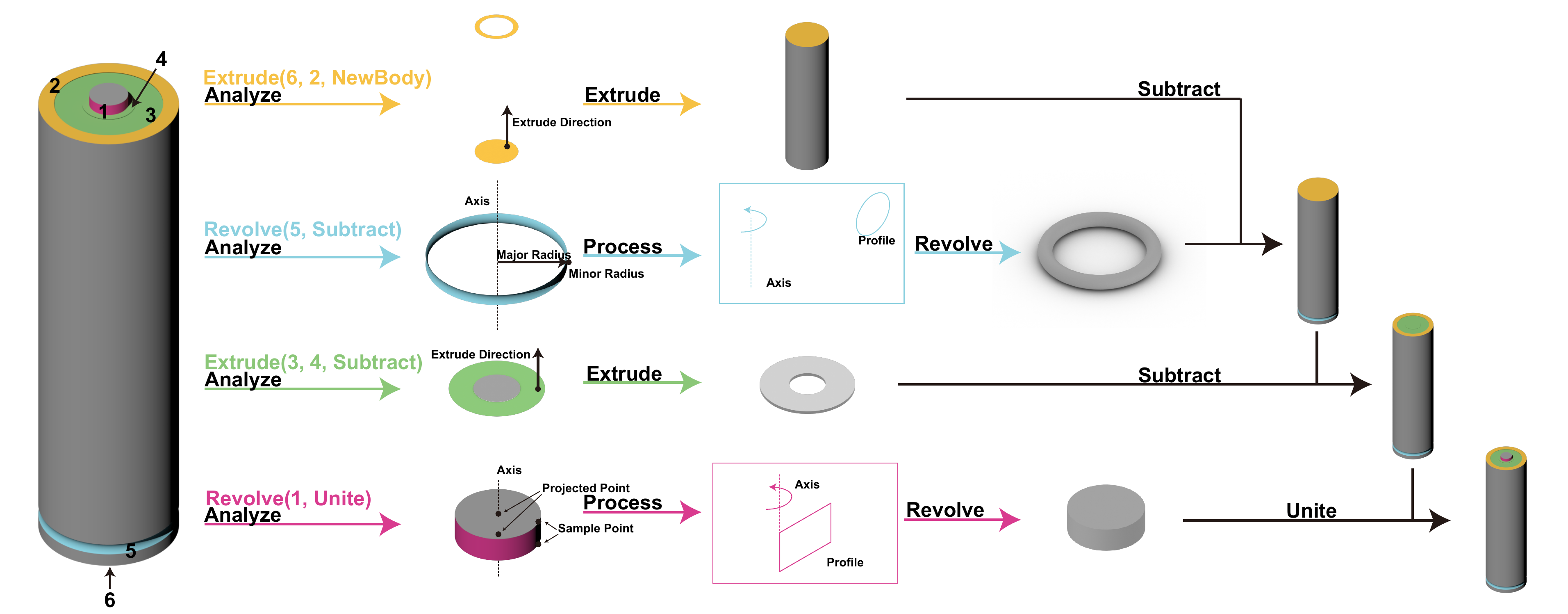} 
  \caption{Battery modeling process using the gym interface. The leftmost image shows face IDs on the surfaces. The right sequence illustrates four extrusion/revolution operations with Boolean operations applied iteratively to generate the final geometry.} 
  \label{fig:actions} 
\end{figure*}

We develop a CAD training gym based on the Parasolid geometric engine. By encapsulating the API of the CAD engine, we design a set of high-level development interfaces for command-sequence-based modeling. Each interface corresponds to an extrusion or revolution operation combined with a Boolean operation (newbody, intersection, union, or subtraction) that incrementally modifies the geometry. The gym also supports parallel sampling across multiple scenarios, improving RL training efficiency.

To illustrate the modeling process, we take battery construction as an example. As shown in Figure~\ref{fig:actions}, the initial B-Rep geometry is represented as a face-adjacency graph with unique face IDs. The gym defines two fundamental modeling operations, allowing the RL algorithm to explore different command sequences and parameter settings. The trained model generates command sequences that closely approximate the input B-Rep.

Our Gym is encapsulated into the following domain-specific language (DSL):
\begin{align*}
&M:=G;[X] \\
&X:=E\ |\ R \\
&E := \text{add\_extrude}(F,F,O) \\
&R := \text{add\_revolve}(F, O) \\
&F := \text{face ID} \\
&O := \text{newbody}\ |\ \text{intersection}\ |\ \text{union}\ |\ \text{subtraction}
\end{align*}

Each generated model $M$ can be represented by a current geometric state $G$ and a command sequence $X$ that modifies the current geometric model. The command sequence $X$ includes the extrude operation $E$ and the revolve operation $R$. The extrude operation $E$ takes two face IDs as its starting and ending faces, along with a Boolean operation $O$ that determines whether the extruded face is combined with the original model via newbody, intersection, union, or subtraction. The revolve operation $R$ takes a face ID as the target face, which is geometrically parsed to obtain the rotation angle, axis, and profile. Additionally, a Boolean operation $O$ determines the composition manner with the original model.

\subsection{Face-Extrusion Operation}
\label{sec:face-extrusion}

As shown in Figure~\ref{fig:actions}, the extrusion operation takes a sketch as the initial geometry and then extrudes it from 2D to 3D. In the extrusion process, we need to specify a distance parameter to determine how far the profile is extruded along the normal direction. Additionally, a Boolean operator is required to specify whether the operation is newbody, intersection, union, or subtraction. Thus, our operation can be represented as $\{\text{start face}, \text{end face}, \text{op}\}$. Here, the start face and end face are a set of parallel planes on the target object, and op represents the type of Boolean operation. The start face defines the sketch for the extrusion operation, while the end face marks the extrusion distance. Therefore, the shape of the end face does not need to be identical to the start face; it only needs to be parallel to it. 

\subsection{Revolution Operation}
\label{sec:revolve}

The revolution operation takes a profile as the revolution surface, a line coplanar with the profile as the axis of rotation, and a rotation angle, then rotates the 2D profile around the axis under the rotation angle to form a 3D shape. We select a curved surface on the target geometry as the resulting surface of the revolve operation. The operation can be represented as $\{\text{face},\text{op}\}$, where $\text{face}$ represents the revolution surface, and $\text{op}$ denotes the type of Boolean operation. 

\begin{figure}[H]
  \centering
  \includegraphics[width=0.48\textwidth]{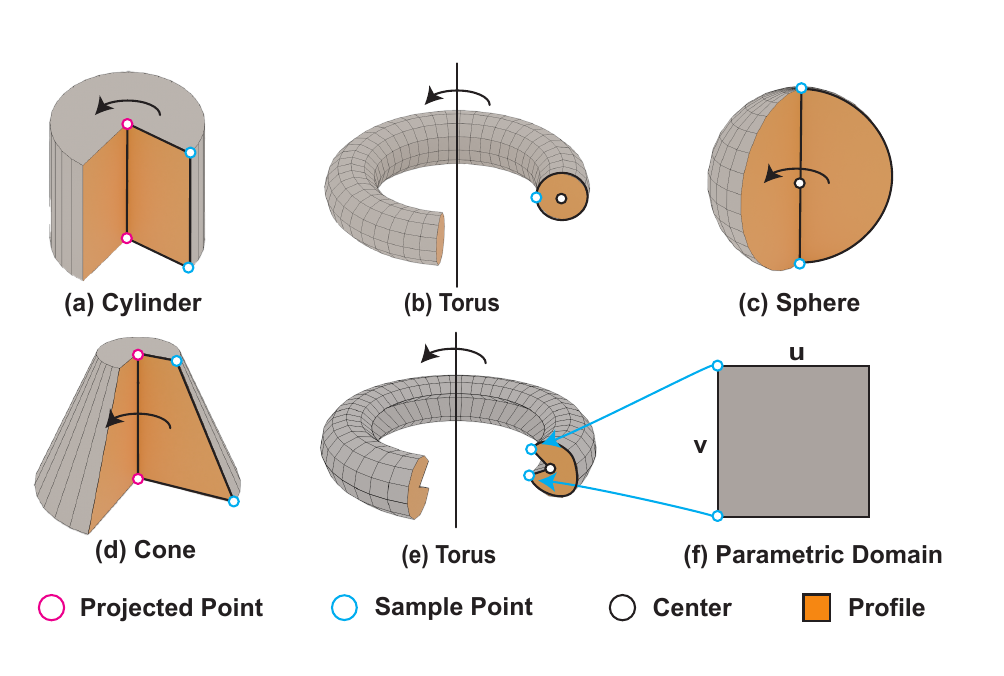} 
  \caption{Revolution operation. The rotation profile is constructed using sampling points from the parametric domain, which are projected and connected based on the surface type. For cylinders and cones, points are projected onto the rotation axis to form the profile. For tori and spheres, the profile arc is determined by the major and minor radii or the sphere center and radius. The rotation angle is defined by the parameter range in the parametric domain.} 
  \label{fig:revolve} 
\end{figure}

We can extract the rotation axis from the B-Rep model and then use feature extraction to calculate the rotation profile and angle. As shown in Figure~\ref{fig:revolve}, we first sample the parametric domain of the given surface to obtain the blue sampling points at $\{u_{min},v_{min}\}$ and $\{u_{min},v_{max}\}$. For cylindrical (Figure~\ref{fig:revolve}a) and conical (Figure~\ref{fig:revolve}d) surfaces, since these surfaces can be generated by revolving a generatrix around an axis, we project the sampling points onto the rotation axis to obtain the red projection points. Connecting these points in order forms the profile shown in the black box, which defines the profile of the revolve operation. In contrast, toroidal and spherical surfaces require a different approach. For a torus, we determine the center using the major radius and construct the profile arc by combining the sampling points with the minor radius. If the profile arc forms a full circle, it is used directly as the profile (Figure~\ref{fig:revolve}b); otherwise, the sampling points are connected to the center to form a partial circular arc (Figure~\ref{fig:revolve}e). For a sphere (Figure~\ref{fig:revolve}c), the center and radius define the profile arc, which is further connected to a line segment passing through the center of the sphere to complete the profile, establishing the revolve profile. The rotation angle is determined by the difference in the range of the parameter $u$ in the parametric domain provided by Parasolid, while the range of the parameter $v$ corresponds to either the length of the generatrix (for cylinders and cones) or the degree of the profile arc (for tori and spheres). 

\section{Generation of CAD Command Sequence}
\label{sec:generation}

In this section, we will delve into the training process of command sequence generation. 
Extensive experiments across different RL frameworks indicate that the Proximal Policy Optimization (PPO) algorithm~\cite{schulman2017proximal} produces the highest-quality command sequences.

\subsection{Proximal Policy Optimization}

PPO is a widely adopted on-policy RL algorithm that supports parallel sampling over multiple environments. Its key advantage is the use of an $\varepsilon$-clipped surrogate loss, which limits the update step between the new and old policies, thereby preventing drastic changes that could destabilize training. The PPO objective is defined as:
\begin{align*}
    L_\text{sur}(\theta) &= ~\mathbb{E}_t \Bigg(\min \bigg( \frac{\pi_\theta(a_t|s_t)}{\pi_{\theta_\text{old}}(a_t|s_t)}\widehat A_t, \\
    &\mathrm{clip}\Big(\frac{\pi_\theta(a_t|s_t)}{\pi_{\theta_\text{old}}(a_t|s_t)},1-\varepsilon,1+\varepsilon\Big)\widehat A_t \bigg)\Bigg),
\end{align*}
where $\widehat A_t$ is computed via Generalized Advantage Estimation (GAE), which combines the advantages of Monte Carlo estimation and Temporal Difference (TD) methods, effectively balancing TD error and variance~\cite{SMLJA16}:
\begin{align*}
    \widehat A_t = \sum_{k=0}^{|\tau|-t}(\gamma\lambda)^{k}\delta_{t+k},
\end{align*}
where $\delta_t = r_t+\gamma V_{\eta_\text{old}}(s_{t+1})-V_{\eta_\text{old}}(s_t)$ and $\lambda\in(0,1)$ is a discount factor. 

The value network $V_\eta$ following policy $\pi$ is updated by minimizing an $\varepsilon$-clipped value loss:

\begin{align*}
L_\text{val}(\eta) &= ~ \mathbb{E}_t
\Bigg(\max\bigg(\mathrm{clip}\Big(V_{\eta}(s_t),V_{\eta_\text{old}}(s_t)-\varepsilon,\\
&V_{\eta_\text{old}}(s_t)+\varepsilon)-\widehat R_t\Big)^2,\Big(V_{\eta}(s_t)-\widehat R_t\Big)^2\bigg)\Bigg)
\end{align*}
with $\widehat R_t = \widehat A_t+V_{\eta_\text{old}}(s_t)$.

\subsection{Face Adjacency Graph}
We systematically convert B-Rep models into attributed graphs through three sequential stages: surface feature extraction, adjacency construction, and graph normalization. Let \( G = (\mathcal{V}, \mathcal{E}) \) denotes the graph where nodes \( v_i \in \mathcal{V} \) represent CAD faces and edges \( e_{ij} \in \mathcal{E} \) encode face adjacency~\cite{jayaraman2021uv}.  

Each node aggregates surface attributes to capture local geometry. The surface type is encoded as an 8-dimensional one-hot vector \( \mathbf{x}^{(i)}_\text{type} \), where \( \mathbf{x}^{(i)}_\text{type}(k) = 1 \) if and only if \( v_i \) belongs to the \( k \)-th predefined surface category (e.g., Plane, Cylinder). To ensure scale invariance, 100 sample points \( \{\mathbf{p}_k\} \) on the face are normalized by axis-aligned bounding box diagonal \( \Delta = \max(\Delta_x, \Delta_y, \Delta_z, 10^{-2}) \), resulting in a feature \( \mathbf{x}^{(i)}_\text{points} = \left[ \frac{\mathbf{p}_1}{\Delta}, \dots, \frac{\mathbf{p}_{100}}{\Delta} \right] \in \mathbb{R}^{300} \). Normal vectors \( \{\mathbf{n}_k\} \) at these points are concatenated into \( \mathbf{x}^{(i)}_\text{normals} \in \mathbb{R}^{300} \), while a binary trimming mask \( \mathbf{x}^{(i)}_\text{mask} \in \{0,1\}^{100} \) indicates parametric validity, with \( \mathbf{x}^{(i)}_\text{mask}(k) = 1 \) if \( \mathbf{p}_k \) lies within the valid trimmed region. The final node feature combines these attributes as \( \mathbf{X} = [\mathbf{x}_\text{type}, \mathbf{x}_\text{points}, \mathbf{x}_\text{normals}, \mathbf{x}_\text{mask}] \in \mathbb{R}^{708} \).  

Edges are derived from B-Rep topology links to encode global structure. For each directional link \( l \in \mathcal{L} \), bidirectional edges are created by validating node existence: \( \mathcal{E} = \left\{ (i,j) \,|\, \exists \, l: v_i = \text{source}(l), \, v_j = \text{target}(l), \, v_i, v_j \in \mathcal{V} \right\} \cup \left\{ (j,i) \,|\, (i,j) \in \mathcal{E} \right\} \). The adjacency matrix \( \mathbf{A} \in \{0,1\}^{|\mathcal{V}| \times |\mathcal{V}|} \) is symmetrized through \( \mathbf{A} = \max(\mathbf{A}_\text{raw}, \mathbf{A}_\text{raw}^\top) \), where \( \mathbf{A}_\text{raw} \) represents the initial directional adjacency.  

Graph normalization stabilizes training by augmenting self-loops and applying row-wise scaling. Self-connections are added via the identity matrix \( \mathbf{I} \), and degree normalization is performed using the degree matrix \( \mathbf{D} \), where \( D_{ii} = \sum_j (A_{ij} + \delta_{ij}) \) with \( \delta_{ij} \) as the Kronecker delta~\cite{kozen2007indefinite}. The normalized adjacency is computed as \( \widehat{\mathbf{A}} = \mathbf{D}^{-1}(\mathbf{A} + \mathbf{I}) \), ensuring \( \forall i , \sum_j \widehat{A}_{ij} = 1 \, \). Isolated nodes are handled by zeroing invalid degrees, and the result is stored in sparse coordinate format for computational efficiency.  

\subsection{Action Encoding}

In CAD modeling tasks, efficient action representation is crucial for the performance of RL algorithms. We propose an index mapping method that directly maps CAD operations to a discrete action space, effectively reducing the dimensionality of the state-action space. Each CAD operation is defined by four key components: $a = (f_s, f_e, o_t, a_t)$, where $o_t$ represents the operation type (including newbody, intersection, union, and subtraction),  $a_t$ indicates the action type (extrude and revolve), for extrude operation $f_s$ denotes the starting face ID, $f_e$ denotes the ending face ID; for revolve operation $f_s$ equals $f_e$. 

We observe that not all actions are valid, and invalid actions can interfere with the agent's learning process, making the network difficult to converge. To address this issue, we perform a validity check on the actions to identify all valid actions. The number of valid actions is also used to define the action space. We denote the action space as $\mathcal{A}$, and set $\mathcal{A}=\mathcal{A}_{\text{valid}}$. The valid action generation process is described in Algorithm~\ref {alg:valid_action}, which first identifies planar and non-planar faces in the target graph, and then groups planar faces based on their normal vectors. For each group of planar faces and non-planar faces, possible extrusion and revolution operations are generated. To rapidly determine feasibility, each candidate operation is executed individually within the CAD training gym, leveraging geometric engine constraints to validate whether the operation can be successfully performed.

\begin{algorithm}
\caption{Valid Action Generation Algorithm}
\label{alg:valid_action}
\begin{algorithmic}[1]
\State Partition $V$ into planar $P$, non-planar $S$
\For{$g \gets \text{GroupParallelFaces}(P)$} 
    \ForAll{$p_i \neq p_j \in g$}
        \State $\mathcal{A}_{\text{valid}} \gets \mathcal{A}_{\text{valid}} \cup \{\text{ValidExtrude}(p_i,p_j)\}$ 
    \EndFor
\EndFor
\For{$s \in S$}
    \State $\mathcal{A}_{\text{valid}} \gets \mathcal{A}_{\text{valid}} \cup \{\text{ValidRevolve}(s)\}$
\EndFor
\State \Return $\mathcal{A}_{\text{valid}}$
\end{algorithmic}
\end{algorithm}

\subsection{Policy Network}

Our policy network consists of feature extraction modules and a Critic-Actor network. As shown in Figure~\ref {fig:pipeline}, the network processes three inputs: B-Reps of the target and current models, and historical action sequences.
First, we extract B-Rep modeling of the target and current models, encoding each B-Rep graph through an independent UV-Net to produce 256-dimensional embeddings \(g_t\) (target) and \(g_c\) (current). These embeddings are stacked into \(G_{stack} \in \mathbb{R}^{2\times256}\), which undergoes 8-head self-attention alignment, where \(G_{stack}\) serves as both Query and Key/Value
inputs. This generates the refined tensor \(\tilde{G}_{stack}\). We then split \(\tilde{G}_{stack}\) along the stacking dimension to obtain enhanced features \(\tilde{g}_t\) and \(\tilde{g}_c\), finally concatenating them into a fused geometric representation \(G_{cat} \in \mathbb{R}^{512}\).

For action sequence processing, historical actions \((a_0, a_1, \dots, a_t)\) are first embedded into 256-dimensional vectors with sinusoidal positional encodings, then fed into an 8-head Gated Transformer-XL (GTrXL)~\cite{parisotto2020stabilizing} module composed of \(L\) stacked blocks. Each GTrXL block stabilizes the modeling of long-range dependencies through a gating mechanism. Finally, we extract the temporal features  \(h_{\text{last\_valid}}^{(L)}\) from the hidden state of the last valid action in the \(L\)-th layer. We empirically choose the number of stacked layers \(L = 3\). 

Cross-modal fusion aligns geometric and action features through attention mechanisms: \(h_{\text{last\_valid}}^{(L)}\) serves as Query while stacked geometric features \(\tilde{G}_{stack}\) act as Key/Value, producing action-conditioned context \(F_a \in \mathbb{R}^{256}\). The fused features \(F_a\) and \(G_{cat}\) are concatenated into a 768-dimensional vector, which is projected to 2048 dimensions through an MLP.

Finally, the Critic-Actor network processes the 2048-dimensional feature using a dual-head MLP to estimate the cumulative rewards and action probabilities. We use dropout layers to avoid overfitting and ReLU activations to preserve non-linearity.

\subsection{Reward Shaping}  
\label{sec:reward}
The reward function plays a critical role in guiding the agent’s exploration during the CAD modeling process. We begin by adopting Intersection over Union (IoU) as the foundational metric for measuring global volumetric alignment between the generated model \(\mathcal{G}\) and the reference model \(\mathcal{S}\).

\textbf{Intersection over Union (IoU)} is utilized to measure the similarity between generated models and the ground truth.
\begin{equation*}
    \mathrm{IoU}(\mathcal{G}, \mathcal{S}) = \frac{\mathcal{G} \cap \mathcal{S}}{\mathcal{G} \cup \mathcal{S}},  
\end{equation*}  
where \(\mathcal{G} \cap \mathcal{S}\) denotes the overlapping volume between the reference and generated models, and \(\mathcal{G} \cup \mathcal{S}\) represents their combined volumetric union. A value of 1 indicates perfect alignment, while 0 signifies no overlap.  

However, observations reveal that relying solely on IoU led to suboptimal policies. The agent prioritized maximizing coarse volumetric overlap while neglecting fine geometric details. To address this limitation, we introduce three complementary rewards: Minimum Matching Distance (MMD)~\cite{achlioptas2018learning}, Normal Consistency (NC)~\cite{yu2022monosdf}, and Neural Reward (NR), which explicitly enforce local geometric fidelity, surface quality, and high-dimensional feature similarity.

\textbf{Minimum Matching Distance (MMD)} quantifies the average distance between the generated model and its closest-matching reference shape. It leverages Chamfer Distance (CD) and Earth Mover’s Distance (EMD) \cite{achlioptas2018learning} to measure bidirectional geometric discrepancies. For two point clouds \(\mathcal{X} = \{x_i\}_{i=1}^N\) and \(\mathcal{Y} = \{y_j\}_{j=1}^M\)sampled from the surface:

\textbf{ Chamfer Distance (CD)} calculates point-wise proximity between point clouds sampled from \(\mathcal{X}\) and \(\mathcal{Y}\):  
\begin{equation*}
\mathrm{d}_{CD}(\mathcal{X}, \mathcal{Y}) = \frac{1}{|\mathcal{X}|} \sum_{x \in \mathcal{X}} \min_{y \in \mathcal{Y}} \| x - y \|_2^2 + \frac{1}{|\mathcal{Y}|} \sum_{y \in \mathcal{Y}} \min_{x \in \mathcal{X}} \| y - x \|_2^2.  
\end{equation*}

\textbf{Earth Mover’s Distance (EMD)} measures the dissimilarity between two point clouds more accurately by finding the optimal bijection. It is widely used to evaluate the performance of point cloud reconstruction and completion with excellent results.

\begin{equation*}
    \mathrm{d}_{EMD}(\mathcal{X}, \mathcal{Y}) = \min_{\phi: \mathcal{X} \to \mathcal{Y}} \sum_{x \in \mathcal{X}} \| x - \phi(x) \|_2,  
\end{equation*}
where $\phi$ is a bijection.

The MMD metric aggregates the minimum CD and EMD values between each reference shape and all generated candidates:  

\begin{equation*}
    \text{MMD-CD}(\mathcal{G}, \mathcal{S}) = \frac{1}{|\mathcal{S}|} \sum_{\mathcal{Y}\in \mathcal{S}} \min_{\mathcal{X} \in \mathcal{G}} d_{CD}(\mathcal{X}, \mathcal{Y}),  
\end{equation*}
\begin{equation*}
    \text{MMD-EMD}(\mathcal{G}, \mathcal{S}) = \frac{1}{|\mathcal{S}|} \sum_{\mathcal{X} \in \mathcal{G}} \min_{\mathcal{Y} \in \mathcal{S}} d_{EMD}(\mathcal{X}, \mathcal{Y}).  
\end{equation*}

The MMD reward term is formulated as:  

\begin{equation*}
    \mathrm{MMD} = -\frac{1}{2} \left( \text{MMD-CD}(\mathcal{G}, \mathcal{S}) + \text{MMD-EMD}(\mathcal{G}, \mathcal{S}) \right),  
\end{equation*}
where the negative sign converts distance minimization into reward maximization.  

\textbf{Normal Consistency (NC)} explicitly rewards alignment of surface normals, critical for preserving sharp edges and smooth curvature transitions. Let \(n_s\) and \(n_g\) denote the normal vectors of points \(s \in \mathcal{S}\) and \(g \in \mathcal{G}\), respectively. NC measures the maximum cosine similarity between corresponding normals:  

\begin{equation*}
    \mathrm{NC}(\mathcal{G}, \mathcal{S}) = \frac{1}{|\mathcal{S}|} \sum_{s \in \mathcal{S}} \max_{g \in \mathcal{G}} \left( n_s \cdot n_g \right).  
\end{equation*}

\textbf{Neural Reward (NR)} addresses the limitations of conventional geometric rewards, which have limited sensitivity to subtle CAD features. NR leverages UV-Net’s B-Rep embeddings: by reformulating rewards as cosine similarities between these high-dimensional features, it enables better feature-aware optimization. It is formulated as:

\begin{equation*}
    \mathrm{NR}(\mathcal{G}, \mathcal{S}) = \cos\left(f_{UV-Net}\left(\mathcal{G}\right ), f_{UV-Net}\left(\mathcal{S}\right)\right ).  
\end{equation*}

The final composite reward function combines these metrics through a weighted sum:

\begin{align*}
    \mathrm{R} &= ~ \alpha \cdot \text{IoU}(\mathcal{G}, \mathcal{S}) + \beta \cdot \text{MMD} 
    + \gamma \cdot \text{NC}(\mathcal{G}, \mathcal{S})\\
    &+ \delta \cdot \text{NR}(\mathcal{G}, \mathcal{S}).
\end{align*}  
where the weights \(\alpha = 0.3\), \(\beta = 0.2\), \(\gamma = 0.2\), and \(\delta = 0.3\) are determined empirically to balance global shape alignment (IoU), local geometric fidelity (MMD), surface quality (NC), and semantic similarity (NR).
\section{Experiments and Comparison}
\label{sec:experiment}


We evaluate our CAD command sequence generation algorithm on a workstation equipped with an NVIDIA L20 GPU (48GB memory).

\paragraph{\textbf{Datasets}}



\LXY{We construct our datasets from two primary sources: 170$k$ models from the DeepCAD dataset and a custom-curated corpus of 20$k$ models drawn from the ABC dataset~\cite{koch2019abc}. To construct this custom corpus, we first conducted online parsing and deduplication, retaining only models involving extrusion and revolution operations. From this filtered pool, we then sampled 20$k$ models using random stratified sampling based on geometric complexity categories: simple (fewer than 10 faces), medium (10–20 faces), and complex (more than 20 faces).}

\LXY{From the custom corpus of 20$k$ models, we allocated a 1$k$-model test dataset for final evaluation using a random stratified sampling strategy, where 39.8\% of models contain more than 20 faces. The remaining 19$k$ models were combined with the 170$k$ DeepCAD models to form a 189$k$-model pre-training dataset. This pre-training dataset covers a wide range of geometric complexities: $33\%$ of models have fewer than 10 faces, $32\%$ have 10–20 faces, and $35\%$ have more than 20 faces.}

\subsection{Evaluation Metrics}
To assess the quality of the generated CAD models, we first convert the B-Rep models into point clouds. Let $\mathcal{S}$ denote the reference models and $\mathcal{G}$ represent the generated models. 
We employ IoU, COV, MMD-CD, JSD, and NC as metrics to evaluate the quality of our generated CAD models~\cite{achlioptas2018learning}. IoU, MMD-CD, and NC have been introduced in Section~\ref{sec:reward}.

\textbf{Coverage (COV)} evaluates the quality of 3D shape synthesis by measuring whether the generated shape set covers the reference shape set. Specifically, the COV metric represents the proportion of shapes in the reference set that are matched to at least one shape in the generated set. If every shape in the reference set can be matched to at least one shape in the generated set, the COV value is 1. Otherwise, if some shapes in the reference set cannot be matched, the COV value is less than 1.

\begin{equation*}
    \text{COV}(\mathcal{G},\mathcal{S}) = \frac{|\{\arg \min_{\mathcal{Y}\in \mathcal{S}}d_{CD}(\mathcal{X},\mathcal{Y})|\mathcal{X}\in\mathcal{G}\}|}{|\mathcal{S}|}.
\end{equation*}

\textbf{Jensen-Shannon Divergence (JSD)} measures the dissimilarity between two point clouds from the perspective of voxel distribution:
\begin{equation*}
    \text{JSD}(P_{\mathcal{G}},P_{\mathcal{S}}) = \frac{1}{2}D(P_{\mathcal{S}}||M) + \frac{1}{2}D(P_{\mathcal{G}}||M),
\end{equation*}
where $M = \frac{1}{2}(P_{\mathcal{S}} + P_{\mathcal{G}})$ and D is the KL-divergence. $P_{\mathcal{G}}$ and $P_{\mathcal{S}}$ are distributions of points in the generated and reference models.

\subsection{Training Efficiency Comparison}


Fusion 360 provides a server-side plugin called Fusion 360 Gym~\cite{willis2021fusion}, which sequentially receives CAD modeling commands sent from external programs and returns the results to those programs via Hypertext Transfer Protocol (HTTP). Based on this plugin, we connect our reinforcement learning algorithm with Fusion 360 to enable closed-loop training.

We compared our novel training gym with Fusion 360 Gym over 10,000 training steps. Using parallel computation (1, 8, and 16 environments), we evaluated both memory consumption and execution time per step.
Table~\ref {tab:memory} shows that Fusion 360 Gym consumes substantially more memory. The frequent use of the revert operation in Fusion 360 introduces a risk of memory overflow. In contrast, our gym employs a stable mark-and-revert mechanism that maintains steady memory usage without significant fluctuations. In addition, by replacing Fusion 360's network communication mechanism with multi-process parallelism and shared memory architecture, we achieve a significant reduction in data transmission latency between the reinforcement learning algorithm and the gym. We get up to $39X$ speedup compared with Fusion 360 Gym, as shown in Table~\ref{tab:speedup}. \YXL{More discussion of the limitations of Fusion 360 Gym can be found in  ~\ref{appendix-1}.}

\begin{table}[htbp]
    \centering
    \caption{Comparison of memory usage between Fusion 360 and our gym  (unit: MB). It shows that Fusion 360 Gym consumes substantially more memory, while our training gym maintains steady memory usage.}
    \label{tab:memory}
    \resizebox{\linewidth}{!}{
    \begin{tabular}{cp{1.5cm}p{1.5cm}p{1.5cm}p{1.5cm}}
        \hline
        & Fusion & 1-env & 8-env &  16-env \\
        \hline
         Initial Memory & 1203.6  & 9.3  & 10.4 & 9.5  \\
        \hline
         Final Memory & 6362.7  & 10.2  & 11.8 & 10.3  \\
        \hline
    \end{tabular}
    }
\end{table}

\begin{table}[htbp]
    \centering
    \caption{Comparison of execution time per step (unit: ms). It shows that our method gets up to $39X$ speedup compared with Fusion 360 Gym.}
    \label{tab:speedup}
    \resizebox{\linewidth}{!}{
    \begin{tabular}{cp{1.5cm}p{1.5cm}p{1.5cm}p{1.5cm}}
        \hline
        & Fusion & 1-env & 8-env & 16-env \\
        \hline
        Execution Time   & 1674  & 198  & 52.5 &  42.18 \\
        \hline
        Speedup  & -  & 8.45  & 31.89 & 39.69  \\
        \hline
    \end{tabular}
    }
\end{table}

\subsection{B-Rep Modeling Comparison}

\begin{figure}[tb]
  \centering
  \includegraphics[width=0.9\linewidth]{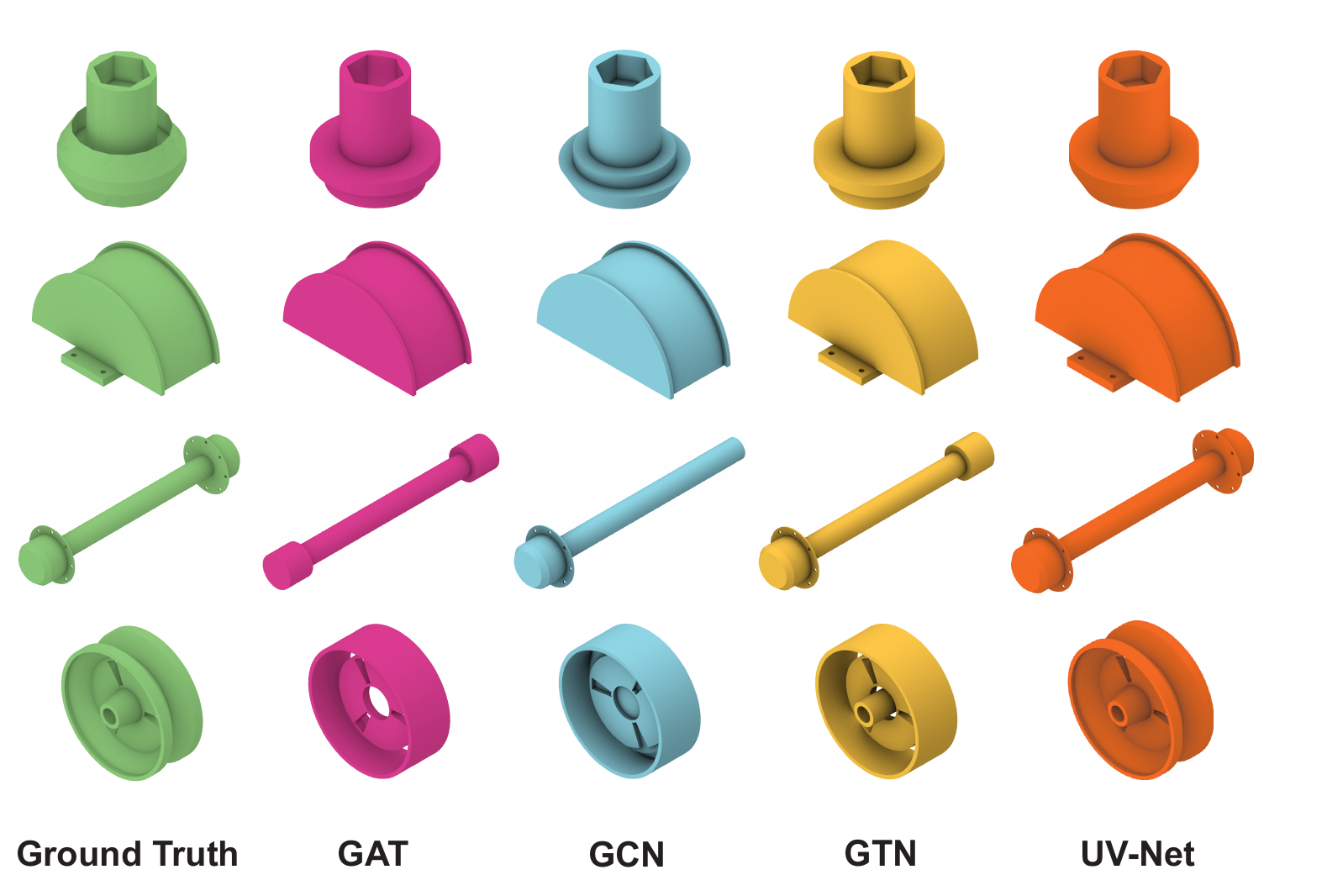}
  \caption{Comparison of different Graph Neural Network (GNN) architectures and UV-Net on reconstruction tasks. It indicates that the UV-Net outperforms the Graph Convolutional Network (GCN), Graph Attention Network (GAT), and Graph Transformer Network (GTN) in reconstructing fine geometric details.}
  \label{fig:gnn} 
\end{figure}

We compare three graph neural network (GNN) architectures and UV-Net for reconstructing B-Rep models:
\begin{itemize}
    \item \textbf{Graph Convolution Network (GCN):} Use fixed-weight neighborhood aggregation, which is computationally efficient but struggles with capturing heterogeneous geometric details.
    \item \textbf{Graph Attention Network (GAT):} Incorporate learnable attention mechanisms; however, its single-head attention limits multi-scale feature integration.
    \item \textbf{Graph Transformer Network (GTN)~\cite{shi2020masked}}: Utilize multi-head self-attention with edge-aware positional encoding, effectively capturing long-range dependencies and global topological constraints.
    \item \textbf{UV-Net:} 
    Exploit the UV parametric domain of curves and surfaces for geometry modeling and adjacency graphs for explicit topological representation. This approach effectively combines convolutional neural networks and graph neural networks to significantly enhance representation capability.
\end{itemize}


To enhance the discriminative power of the ablation analysis, we curated $0.5k$ complex models from the test set to construct the ablated dataset. As shown in Table~\ref{tab:gnn_comparison}, UV-Net outperforms GCN, GAT, and GTN across all five evaluation metrics on this dataset. In particular, it achieves a significant improvement in IoU compared to the previous best-performing method, GTN. The model comparison in Figure~\ref{fig:gnn} further substantiates this finding, demonstrating UV-Net's superior capability in learning latent feature representations of B-Rep models, which results in a closer fit to the ground truth.

\begin{table}[htbp]
    \centering
    \caption{
    Performance comparison of GNN architectures and UV-Net on reconstruction tasks. The direction of the arrows indicates the direction of better performance. It shows that UV-Net outperforms GCN, GAT, and GTN across all five evaluation metrics.
    }
    \label{tab:gnn_comparison}
    \resizebox{\linewidth}{!}{
    \begin{tabular}{cccccc}
        \hline
        & IoU$\uparrow$ & COV$\uparrow$ & MMD-CD$\downarrow$ & JSD$\downarrow$ & NC$\uparrow$ \\
        \hline
        GCN & 0.7291 & 0.5951 & 0.0589 & 0.3232 & 0.7570 \\
        \hline
        GAT & 0.7102 & 0.5941 & 0.0215 & 0.2947 & 0.7573 \\
        \hline
        GTN & 0.7326 & 0.6380 & 0.0510 & 0.2675 & 0.7753 \\
        \hline
        UV-Net & \textbf{0.8207} & \textbf{0.7069} & \textbf{0.0104} & \textbf{0.2303} & \textbf{0.7839} \\
        \hline
    \end{tabular}
    }
\end{table}

\subsection{Training strategy}
Our training strategy consists of two phases: feature extraction and command prediction. Since the command encoding retains only face IDs (without explicit geometric details), the pre-trained command generation network loses effectiveness on new geometries. Therefore, we reinitialize the command generation network for each new input. For the feature extraction network, we perform self-supervised pre-training on \(189k\) CAD models and employ curriculum learning to enhance generalization. We first rank the CAD models in order of increasing complexity based on the number of faces, and then train them sequentially from low to high complexity, updating the model weights along the way.
A comparison with a case-by-case training strategy (where each geometry is trained independently) confirms that curriculum learning leads to superior overall performance, as summarized in Table~\ref{tab:curriculum_learning}.

\begin{table}[htbp]
    \centering
    \caption{
    Comparison of training strategies on our dataset. The curriculum learning-based strategy outperforms the case-by-case approach across all evaluation metrics.
    }
    \label{tab:curriculum_learning}
    \resizebox{\linewidth}{!}{
    \begin{tabular}{cccccc}
        \hline
        & IoU$\uparrow$ & COV$\uparrow$ & MMD-CD$\downarrow$ & JSD$\downarrow$ & NC$\uparrow$ \\
        \hline
        Case by case & 0.8142 & 0.7567 & 0.0194 & 0.2437 & 0.7754 \\
        \hline
        Curriculum & \textbf{0.8624} & \textbf{0.7725} & \textbf{0.0078} & \textbf{0.2076} & \textbf{0.7923} \\
        \hline
    \end{tabular}
    }
\end{table}

\subsection{Reward Ablation Study}


\begin{table}[tb]
    \centering
    \caption{
    Reward function ablation study. It shows that introducing NR significantly improves evaluation metrics of IoU, COV, JSD and NC, while the metric of MMD-CD shows no significant degradation. Therefore, we choose the combination of IoU, MMD, NC, and NR as our final reward configuration.}
    \label{tab:reward-ablation}
    \resizebox{\linewidth}{!}{
    \begin{tabular}{cccccc}
        \hline
        & IoU$\uparrow$ & COV$\uparrow$ & MMD-CD$\downarrow$ & JSD$\downarrow$ & NC$\uparrow$ \\
        \hline
        $\mathrm{IoU}$ & 0.7045 & 0.5748 & 0.0204 & 0.3235 & 0.7437 \\
        \hline
        $\mathrm{IoU + MMD}$ & 0.7436 & 0.6474 & 0.0132 & 0.2577 & 0.7790 \\
        \hline
        $\mathrm{IoU + MMD + NC}$ & 0.7932 & 0.7190 & \textbf{0.0084} & 0.2229 & 0.7889 \\
        \hline
        $\mathrm{IoU + MMD + NC + NR}$ & \textbf{0.8712} & \textbf{0.7408} & 0.0106 & \textbf{0.2178} & \textbf{0.7908} \\
        \hline
    \end{tabular}
    }
\end{table}


Leveraging the ablated dataset curated in Section 6.3, we perform ablation studies to systematically analyze the impact of different reward function combinations on experimental results. Table~\ref{tab:reward-ablation} shows the outcomes of these studies. Reward functions such as IoU, MMD, and NC are used to measure geometric similarity. NR is used to capture the cosine similarity of high-dimensional B-Rep features. The experimental results show that introducing NR significantly improves evaluation metrics of IoU, COV, JSD and NC, meanwhile, the metric of MMD-CD shows no significant degradation. Therefore, we choose the combination of IoU, MMD, NC, and NR as our final reward configuration.

\subsection{Result Comparison of Reconstruction}
\label{comparison}



\begin{table}[tb]
    \centering
    \caption{
    \LXY{Quantitative results on the B-Rep based reconstruction task. Our method significantly outperforms Fusion 360 Gallery across all metrics on our test dataset (1$k$ models).}
    }
    \label{tab:metric_comparion}
    \resizebox{\linewidth}{!}{
    \begin{tabular}{cccccc}
        \hline
        & IoU$\uparrow$ & COV$\uparrow$ & MMD-CD$\downarrow$ & JSD$\downarrow$ & NC$\uparrow$ \\
        \hline
        Fusion 360 Gallery & 0.5678 & 0.4352 & 0.4769 & 0.4968 & 0.6299 \\
        \hline
        Ours & \textbf{0.9001} & \textbf{0.8206} & \textbf{0.0180} & \textbf{0.1945} & \textbf{0.8227} \\
        \hline
    \end{tabular}
    }
\end{table}

\begin{table}[tb]
    \centering
    \caption{
    \LXY{Quantitative comparison with cadrille and CAD-Recode on the reconstruction task, evaluated on the complete CAD-Recode validation dataset (1$k$ models). Our method significantly outperforms both baselines across all metrics.}}
    \label{tab:metric_comparion_recode}
    \resizebox{\linewidth}{!}{
    \begin{tabular}{cccccc}
        \hline
        & IoU$\uparrow$ & COV$\uparrow$ & MMD-CD$\downarrow$ & JSD$\downarrow$ & NC$\uparrow$ \\
        \hline
        cadrille & 0.7603 & 0.7260 & 0.0577 & 0.4904 & 0.6001 \\
        \hline
        CAD-Recode & 0.9264 & 0.8736 & 0.0043 & 0.0885 & 0.8205 \\
        \hline
        Ours & \textbf{0.9779} & \textbf{0.9836} & \textbf{0.0021} & \textbf{0.0618} & \textbf{0.8799} \\
        \hline
    \end{tabular}
    }
\end{table}

We select three representative learning based CAD model generation algorithms for comparison.

\begin{itemize}
    \item \textbf{Fusion 360 Gallery~\cite{willis2021fusion}} extracts continuous geometric parameters from B-Rep models and uses imitation learning to generate command sequences, followed by a confidence-based local search for optimization.
    \item \textbf{CAD-Recode~\cite{rukhovich2024cad}} translates a point cloud into Python code by leveraging a relatively small language model as a decoder and combined with a lightweight point cloud projector. The method is trained on a self-constructed, million-scale training dataset composed exclusively of sketch-extrusion operations.
    \item \textbf{cadrille~\cite{kolodiazhnyi2025cadrille}}  proposes a multi-modal CAD reconstruction model that can simultaneously process three input modalities, including point clouds, images, and text. The method adopts a two-stage pipeline: supervised fine-tuning (SFT) on large-scale procedurally generated data, followed by reinforcement learning fine-tuning using online feedback. 
    
\end{itemize}

\begin{figure*}[!h]
 \centering
  \includegraphics[width=1.0\textwidth]{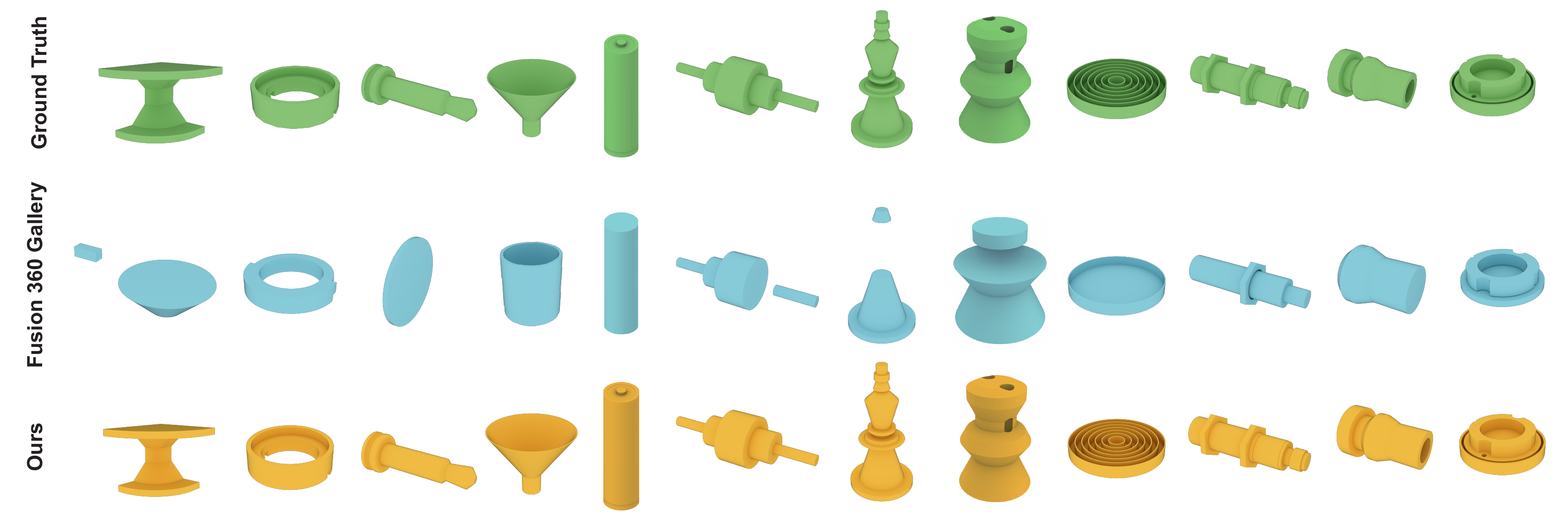} 
  \caption{
  Comparison of generation results with Fusion 360 Gallery. It shows our method generates higher-quality results in terms of both completeness and detail.} 
  \label{fig:models} 
\end{figure*}

\begin{figure}[H]
 \centering
  \includegraphics[width=0.43\textwidth]{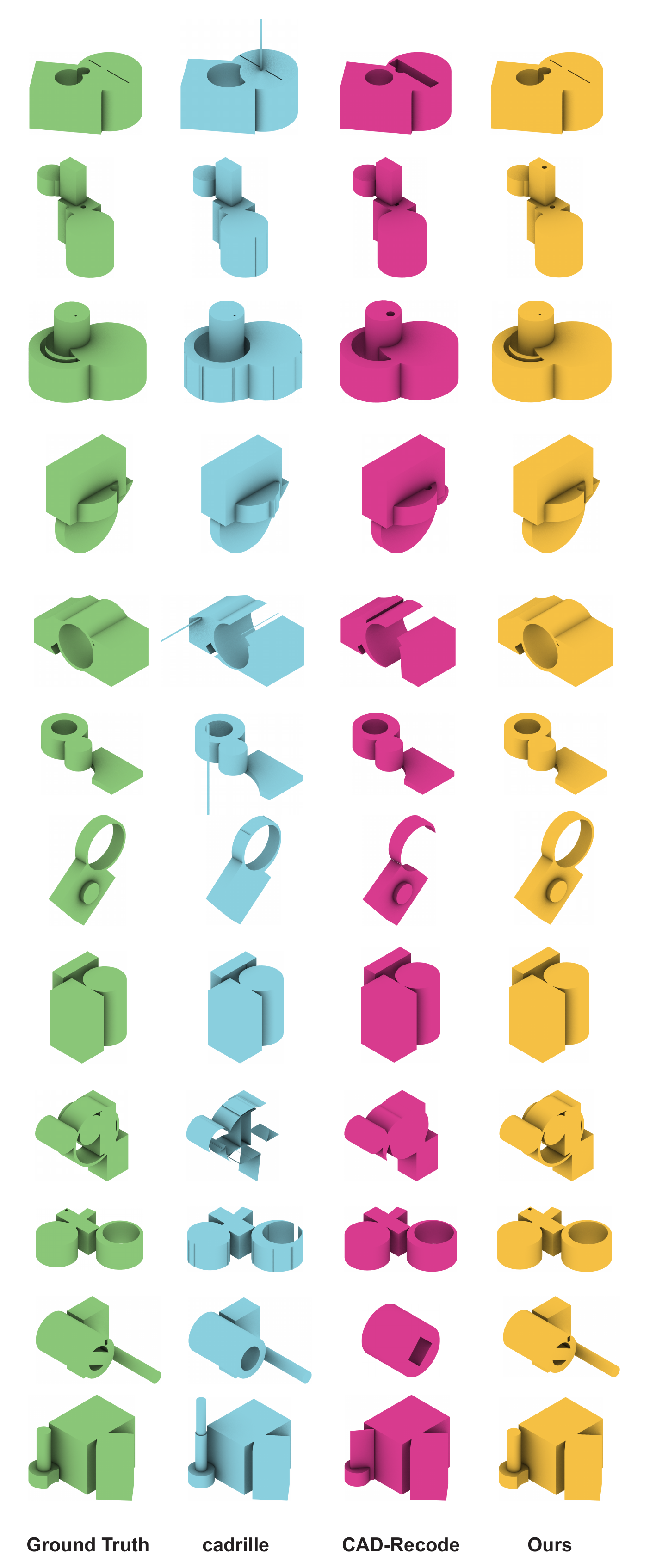} 
  \caption{
Comparison of generation results with cadrille and CAD-Recode. It shows our method generates higher-quality results in terms of both completeness and detail.} 
  \label{fig:models2} 
\end{figure}

Currently, most supervised learning methods are trained on datasets that include only sketch-extrusion operations. To overcome this limitation, we extend the output modules of Fusion 360 Gallery models to support revolve operations; \YXL{more details can be found in  ~\ref{appendix-2}.} And we retrain the model on our dataset. As illustrated in Figure~\ref{fig:models}, our approach outperforms Fusion 360 Gallery in completeness and detail preservation. Quantitative results in Table~\ref{tab:metric_comparion} demonstrate substantial improvements across all evaluation metrics.

To validate the generalization capability of our method on unseen datasets, we select two baseline approaches - CAD-Recode (using point cloud as input) and cadrille (using point cloud and image as input) - both pre-trained on the CAD-Recode dataset, and compare their generation results with our method on the CAD-Recode validation dataset. As illustrated in Figure~\ref{fig:models2} and summarized in Table~\ref{tab:metric_comparion_recode}, which present qualitative results and quantitative metrics respectively, our method consistently outperforms both baselines across all evaluation criteria.

\subsection{\TM{Comparison with Traditional Feature Recognition Methods}}
\begin{figure}[H]
 \centering
  \includegraphics[width=0.45\textwidth]{pic/feature-recogonize.pdf} 
  \caption{\TM{Comparison results with SolidWorks and Autodesk Inventor. The regions highlighted in magenta indicate features that cannot be recognized. We consider four representative cases.}}
\label{fig:compare-feature}
\end{figure}

\TM{To compare with traditional feature-recognition methods, we select the industrial CAD software Dassault SolidWorks~\cite{solidworks} and Autodesk Inventor~\cite{inventor}. Both provide state-of-the-art built-in feature-recognition plugins and can therefore be regarded as representative implementations of advanced feature-recognition techniques. Our experiments show that our method still exhibits several advantages over them, as shown in Figure~\ref {fig:compare-feature}: }

\YXL{(a) After two extrusion operations followed by a Boolean union, traditional feature-recognition methods fail to correctly interpret the new topology and instead recognize a single feature that does not correspond to the original modeling operations. In contrast, our method successfully recovers the two separate extrusion operations.}

\YXL{(b) For extrusion operations whose directions are not perpendicular to the sketch plane, both SolidWorks and Autodesk Inventor fail to recognize the feature (the red region in the figure), whereas our method can handle such cases correctly.}

\YXL{(c) For irregular extrusions and revolutions, traditional feature recognition typically relies on an analytic assumption that the end face can be traced back to a “planar end face + single sketch” configuration. This assumption easily breaks down when the end face is defined as up-to-surface/body/face with an offset, or when the side wall includes variable draft angles, leading to recognition failures. By contrast, our method can still correctly reconstruct the operations even with irregular profiles and non-uniform side walls.}

\YXL{(d) When multiple extrusion or revolution operations are combined using Boolean union/difference/intersection, the resulting model often exhibits complex topology, making it difficult for feature-recognition methods to recover the original “seed” extrusions/revolutions together with their sketches and directions; as a result, the editable feature history is lost. In comparison, our method can still decompose the merged result and reconstruct a plausible and consistent sequence of modeling operations.}




\subsection{Failure Case}
\label{sec:failed}
\begin{figure}[tb]
  \centering
\includegraphics[width=0.5\textwidth]{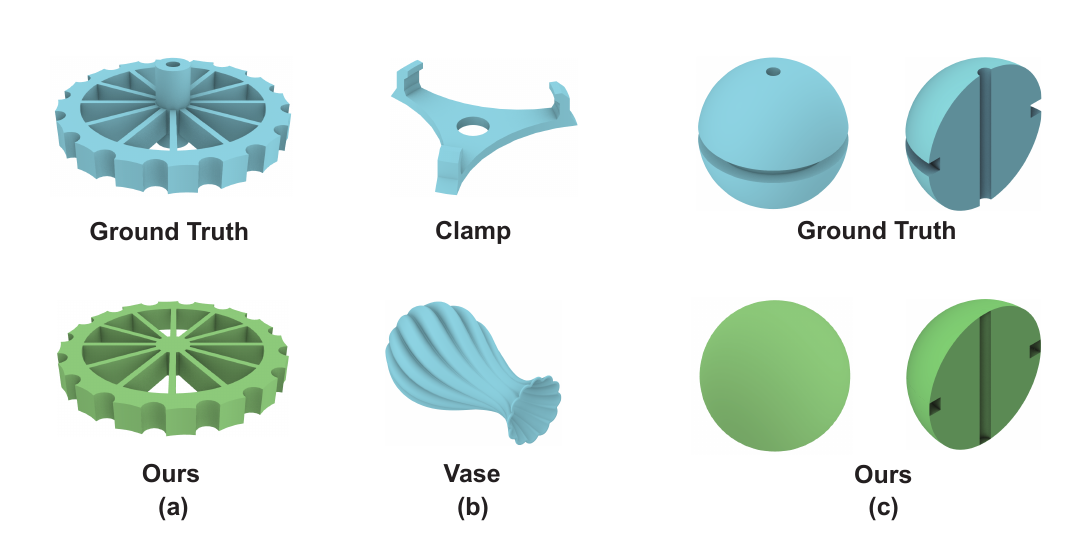} 
  \caption{Examples of failure cases categorized into three types: (a) complex model with many faces; (b) models requiring unsupported operations (e.g., fillet and spline surface); (c) special model trimming.}
  \label{fig:failure} 
\end{figure}

\begin{figure}[tb]
  \centering
  \includegraphics[width=0.45\textwidth]{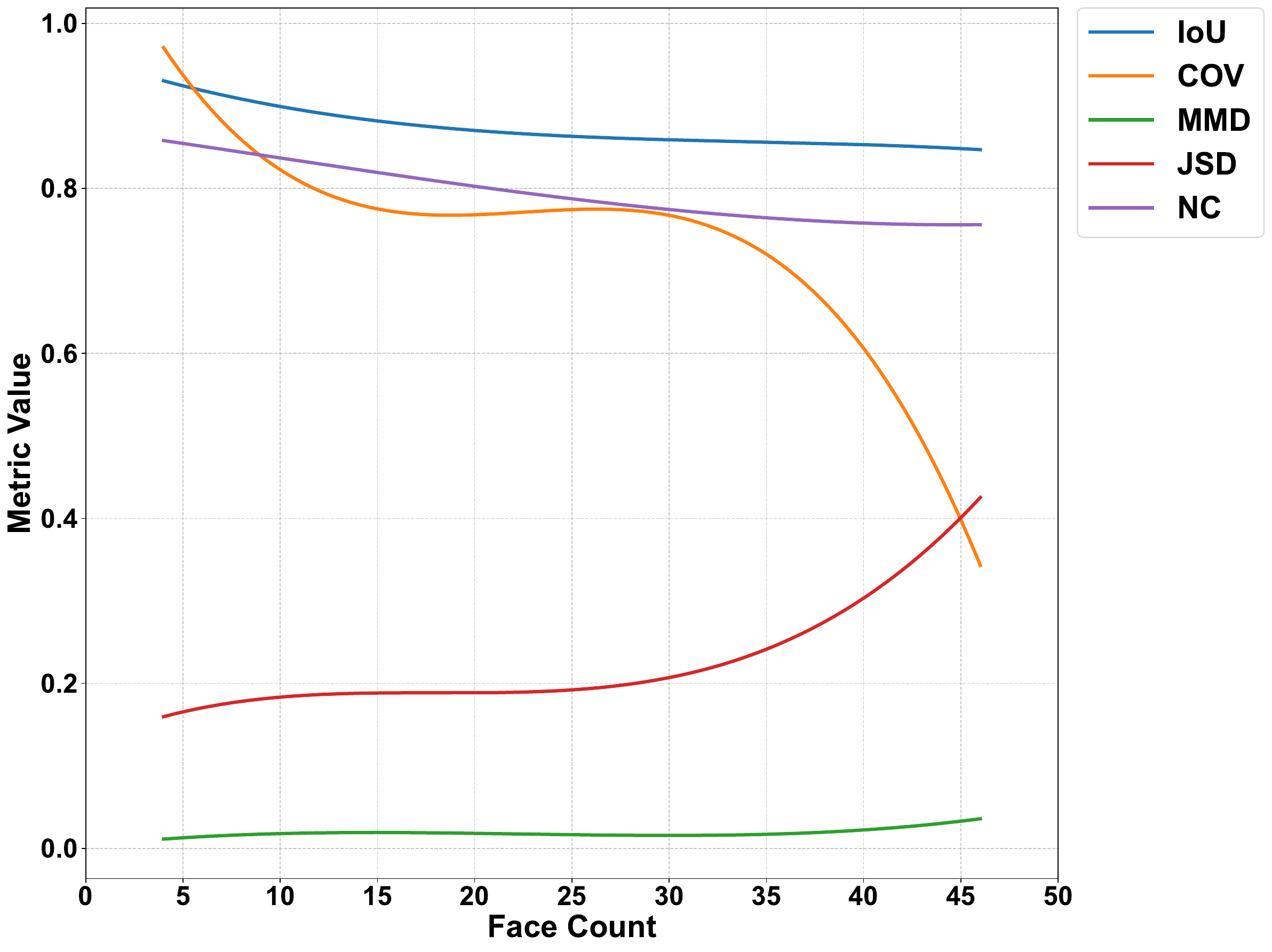} 
  \caption{
The quantitative relationship between model complexity and evaluation metrics. While the global reconstruction quality of our model remains acceptable beyond 35–40 faces, the fidelity of fine-grained geometric details deteriorates noticeably.}
  \label{fig:surface-iou} 
\end{figure}

We categorize the cases that our method currently cannot handle into three types. The first type arises from a complex model with many faces. Figure~\ref{fig:failure}a exhibits the generation result of a complex gear model, in which some details are missing. Figure~\ref {fig:surface-iou} depicts the quantitative relationship between generated model complexity (measured by face count) and various evaluation metrics. For simple models with a low face count, IoU and NC remain relatively high, but gradually decline as complexity increases. COV performs well on models with a low number of faces but declines significantly once the face count exceeds approximately 35–40. JSD, which is initially low, rises rapidly as the face count increases, especially beyond 40 faces. Overall, while the global reconstruction quality of our model remains acceptable beyond 35–40 faces, the fidelity of fine-grained geometric details deteriorates noticeably.

The second type arises from the limitations of the supported operations of our training gym. For example, models with fillet and spline surface, as shown in Figure~\ref{fig:failure}b. We will support such operations in the future.

The third type of failure case arises from the inconsistency between the selected trimmed body and the requirements during model trimming. As shown in Figure~\ref{fig:failure}c, the reconstruction of the ground truth model is completed by clipping a sphere twice: first, a complete sphere is constructed, then a groove based on the extrusion is built at the waist, and a cylinder based on rotation is constructed at the center. Finally, these two entities are cut from the sphere through Boolean operations. But in reality, the outer diameter of the circular ring at the bottom of the waist groove and the height of the rotating surface of the central cylinder are both smaller than the diameter of the sphere. The entities generated by extrusion and rotation are both inside the sphere, so the surface of the sphere remains complete after clipping.

\section{Conclusion and Future Work}
\label{sec:conclusion}
\subsection{Conclusion}

Most current CAD model generation algorithms are based on supervised learning methods, where the validity of the generated geometries cannot be effectively verified and fed back to the training network. We are the first to present a geometric engine-based CAD model training gym and introduce an online RL algorithm based on the gym to CAD model generation. Our RL algorithm has added support for revolution operation, enhancing the capability of generating command sequences for complex models. Experiments demonstrate that our method can significantly improve the precision of the command
sequence generation compared with supervised learning methods.

\subsection{Future Work}

Through extensive experiments, it has been found that our method still has shortcomings in the detailed feature learning of complex models and the supported operation amount. In the next phase, we will focus on developing detail-preserving feature extraction networks that support a broader range of input modalities and expanding the gym to support more commonly used modeling operations.

\section*{References}

\bibliography{mybibfile}

\clearpage

\appendix
\section{\YXL{Appendix}}

\subsection{\YXL{The Difficulty of Fusion 360 Framework Integration in Training Gym}}
\label{appendix-1}


\YXL{Integrating the \emph{revolve} command into Fusion 360 is feasible—as noted in Section~\ref{appendix-2}, we implemented a working plugin and completed the full RL training loop on top of Fusion’s API. However, we ultimately moved away from this setup for \textbf{performance and reliability} reasons: (i) our Parasolid-based environment is \textbf{39$\times$ faster} end-to-end and supports efficient multi-environment parallel sampling and inter-process communication (which are crucial for RL throughput); (ii) Fusion’s plugin model and GUI-bound execution make large-scale, headless parallelization difficult, leading to unacceptable wall-clock training times; and (iii) we observed intermittent memory growth and occasional instability during long-running sessions. For these reasons, while Fusion 360 can express the required operators, the proposed kernel-coupled simulator provides the necessary speed, scalability, and robustness for our experiments.}

\subsection{\YXL{Our Extension to Fusion 360 Gallery}}
\label{appendix-2}

\YXL{We extend the method described in the Fusion 360 Gallery paper to support revolve operations, which requires two components: (i) extending the training network so that it can output revolve operations, and (ii) extending the Fusion 360 plugin to execute revolve commands. This plugin runs inside Fusion 360 and is responsible for executing the modeling commands sent from the client. Both components are required for Fusion 360 to fully support revolve operations in our framework.}

\YXL{More specifically, on the model-output side, we expand the action space from extrude-only to a unified mixed-action mode that supports both extrusion and revolution, while introducing only minimal changes to the inference interface and execution backend. The network now predicts a discrete action type $a \in {\text{extrude}, \text{revolve}}$, a face selection, and a Boolean operation. For extrusion, the prediction remains a pair of parallel, non-coplanar planar faces $(f_s, f_e)$ together with a Boolean operation (newbody/intersection/union/subtraction). For revolution, the output is simplified to a single revolve-eligible face $f_r$ (non-planar, typically cylindrical, conical, spherical, or toroidal) plus a Boolean operation; the revolve axis is not regressed by the network but is instead inferred by the plugin described below.}


\YXL{The Fusion 360 plugin provides extended support for the \emph{revolve} operation in Fusion 360. Given a selected surface of revolution, the system automatically infers a suitable rotation axis and generates a corresponding 2D sketch aligned with that axis. For different types of surfaces of revolution, the plugin adopts specialized construction logic:}

\begin{itemize}
    \item \YXL{\textbf{Cylindrical and conical surfaces:} Feature points are extracted from the boundary in parametric space, projected onto the rotation axis, and then connected in sequence to form the sketch profile.}

    \item \YXL{\textbf{Spherical surfaces:} Boundary points are obtained from the parametric domain and connected to the sphere center to construct the revolve sketch.}
    
    \item \YXL{\textbf{Toroidal surfaces:} If the boundary in parametric space is a complete circle, this circle is directly used as the sketch. Otherwise, boundary points are connected to the center of the minor circle to build the sketch profile.}
\end{itemize}
\YXL{For surface types where a stable rotation axis or sketch cannot be inferred under the above rules, the plugin deliberately rejects processing them to avoid generating unreliable or meaningless features. All generated sketches and features are created within a dedicated reconstruction component to maintain structural consistency. If no valid entity can be found for reconstruction, the corresponding Boolean operations are safely skipped so as not to interfere with the overall modeling workflow.}

\end{document}